\documentclass[11pt]{article}

\usepackage[preprint]{acl}

\usepackage{times}
\usepackage{latexsym}

\usepackage[T1]{fontenc}

\usepackage[utf8]{inputenc}

\usepackage{microtype}

\usepackage{inconsolata}

\usepackage{graphicx}
\usepackage{booktabs}
\usepackage{multirow}
\usepackage{amsmath}
\usepackage[breakable]{tcolorbox}
\usepackage{fancyvrb}
\usepackage{tikz}
\usepackage{pgfplots}
\usepgfplotslibrary{groupplots}
\pgfplotsset{compat=1.18}
\usepackage[table]{xcolor}
\usepackage{enumitem}
\usepackage{fontawesome5} 

\usepackage{xspace}
\newcommand{\name}{\textsc{OmniLingual-MsASR}\xspace}

\title{

\textsc{Omnilingual-MsASR}: Benchmarking Multimodal, Multilingual, Long-Form Multi-Speaker ASR of Omni Large Language Models
}

\author{Yuan Ge$^{1,2}$, Pei Zhang$^{2}$, Xize Cheng$^2$, Lingzhou Mu$^2$, Mingyang Wang$^2$, Jin Xu$^2$, \\
\textbf{Baosong Yang$^2$, Shengxiang Gao$^3$, Jingbo Zhu$^{1,4}$, Zhengtao Yu$^3$, Tong Xiao$^{1,4}$}\thanks{Corresponding author.} \\ 
$^1$ Northeastern University, China \quad
$^2$ Alibaba Token Hub, Alibaba \\
$^3$ Kunming University of Science and Technology \quad
$^4$ NiuTrans Research
}

\begin{document}
\maketitle
\begin{abstract}

Omni-modal large language models (LLMs) have achieved remarkable progress in audio and video understanding. However, existing models struggle with long-form, multi-speaker videos, especially in understanding the entire video and recognizing who says what and when.
To characterize the capability limit of current omni LLMs and guide future model development, we introduce \name, a multilingual, multimodal benchmark for long-form Multi-speaker Automatic Speech Recognition (MsASR) covering six languages, recordings up to 30 minutes, and scenarios with up to 20 speakers. 
\name provides parallel audio and video inputs to evaluate models’ cross-modal consistency and their ability to exploit complementary cues across modalities. 
Extensive evaluations of leading omni-modal LLMs reveal substantial performance bottlenecks, while their tcpWERs frequently exceed $100\%$. These findings highlight fine-grained temporal localization, speaker disambiguation, and cross-modal synergy as key challenges for future omni-modal foundation models
\footnote{Our work is available at \url{https://github.com/QwenLM/Omnilingua-Bench/tree/main/Omnilingua-MSpeaker}.}.

\end{abstract}

\section{Introduction}

\begin{figure}[t]
  \centering
  \includegraphics[width=\columnwidth]{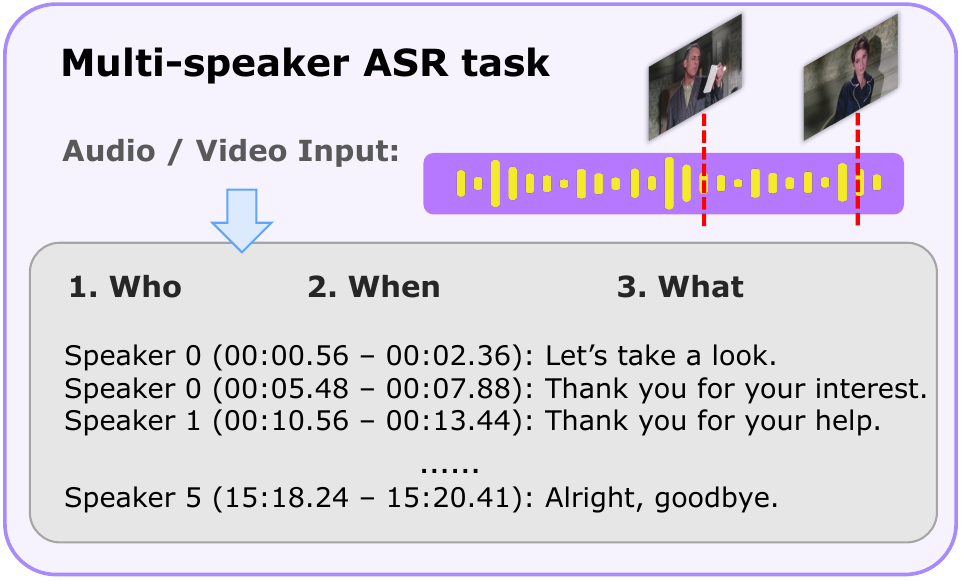}
  \caption{Overview of the MsASR task. The model is required to jointly predicting speaker identities (who), timestamps (when), and text transcriptions (what) in a structured format as shown in Sec. \ref{tab:format}.}
  \label{fig:intro}
\end{figure}

Omni large language models (LLMs) have recently achieved remarkable progress in audio understanding \citep{arora2025landscape, guo2026survey}. Multi-speaker automatic speech recognition (MsASR) reflects this ability by transcribing audio into structured text that specifies who speaks, when each utterance begins and ends, and what is said \citep{he2025survey, wang2024diarizationlm, lin2025diarization, shi2026qwen3}, as shown in Fig. \ref{fig:intro}. Such transcription grounds downstream tasks such as speech translation and voice interaction.

However, existing benchmarks fail to probe the boundary of omni LLMs' audio-visual transcription, where videos in practice are long-form, multilingual, and multi-speaker. 
Traditional MsASR datasets annotate speaker identity, utterance boundaries, and transcripts, but cover only monolingual English or Chinese \citep{kraaij2005ami, fu2021aishell, yu2022m2met, cornell2023chime}. 
Multilingual dialogue benchmarks push forward language coverage, but the audio durations remain far shorter than the context-window limits of modern LLMs \citep{wang2021voxpopuli, sinhamahapatra2026muscat, mu2026summary}. 
Moreover, these benchmarks overlook the potential of the visual modality to aid speaker identification and transcription.

We introduce \textsc{OmniLingual-MsASR}, a multimodal, multilingual, and long-form multi-speaker ASR benchmark that reflects the real audio-visual transcription capability boundary of omni LLMs. 
We collect publicly available videos on diverse topics spanning six languages, namely Spanish, French, Japanese, Korean, Russian, and Thai, with audio durations ranging from short clips of around 20 seconds to long videos of about 30 minutes, each containing up to 20 speakers. 
The MsASR annotations combine an automatic pipeline built from frontier tools with careful manual verification by experienced language experts. The resulting \name comprises roughly 50 videos per language across the six languages, totaling 297 long-form multi-speaker ASR instances.

We conduct extensive evaluations of leading omni-modal LLMs, 
including two open-source models and seven frontier commercial omni LLMs.
Empirical results reveal that while current models achieve robust linguistic transcription, precise temporal grounding remains a major bottleneck: Gemini family omni-LLMs exhibit low cpWER (e.g., 0.23–0.44) but drastically worse tcpWER that frequently exceeds 1.0. Furthermore, model efficacy deteriorates sharply as context scales. Extending audio inputs toward 30 minutes compounds timestamp errors, and scaling speaker counts from two to over seven drives overall speaker confusion from 9.1\% to 25.6\%, causing \texttt{Gemini 3.7 Flash}'s cpWER to surge from 0.387 to 1.113. 
Finally, our parallel audio-video benchmark uncover divergent cross-modal integration strategies across architectures: while visual cues markedly boost temporal localization for \texttt{Doubao-Seed-2.0-Lite}, they act as confounding noise that degrades temporal grounding for \texttt{Gemini 3.7 Flash}. These findings highlight that fine-grained temporal estimation, multi-speaker disambiguation, and cross-modal synergy remain critical for future omni-modal foundation models.
Our contributions are:
\begin{itemize}
    \item We introduce \name, a multimodal, multilingual, and long-form multi-speaker ASR benchmark featuring paralleled audio and video inputs.
    \item We develop a reliable annotation pipeline that integrates audio and visual cues to generate high-quality pseudo-labels, followed by human refinement under standardized guidelines to ensure final data quality.
    \item We conduct a systematic evaluation across widely leading commercial and open-source LLMs, highlighting the current strengths and key challenges in long-form MsASR.
\end{itemize}

\section{Related Work}

\subsection{Omni-modal Large Language Models}

Recent omni-modal large language models extend text-centric LLMs toward unified perception across text, vision, audio, and video, as exemplified by GPT-4o \citep{openai2024gpt4o}, Gemini 2.5 \citep{comanici2025gemini25}, and Qwen3.8-Omni \citep{team2026qwen38}. 
Their audio capabilities have been evaluated on increasingly diverse tasks, including speech recognition, audio question answering, acoustic understanding and reasoning \citep{yang2024airbench, ge2026sagelm, sakshi2025mmau, ge2026emotion}.

However, broad audio competence does not necessarily imply reliable speaker-aware transcription. Recent studies reveal a substantial gap between understanding what was said and identifying who said it \citep{wang2025msubench, kwon2026m3slu}. By requiring models to bind linguistic content to persistent speaker identities across turns and overlaps, speaker-aware transcription provides a concrete, though not exhaustive, indicator of fine-grained conversational understanding.

\subsection{Long-Form Video Understanding}

Long-form video understanding evaluates whether models can retain, retrieve, and reason over information distributed across extended temporal contexts. EgoSchema \citep{mangalam2023egoschema}, LongVideoBench \citep{wu2024longvideobench}, MLVU \citep{zhou2025mlvu}, and Video-MME \citep{fu2025videomme} extend video-language evaluation to minute- or hour-long inputs, with an emphasis on long-range memory, information retrieval, and temporal reasoning.

The emergence of Omni LLMs has further expanded this setting from predominantly visual-language inputs to joint audio-visual perception. LongVALE \citep{geng2025longvale} introduces temporally grounded vision-audio-language events, while OmniVideoBench \citep{li2025omnivideobench}, LVOmniBench \citep{tao2026lvomnibench} and Mula-bench \citep{yang2026mula} evaluate cross-modal reasoning over extended audio-visual contexts \citep{tao2026lvomnibench,yang2026mula}. However, their question-answering and event-description objectives do not capture dense speaker-aware transcription, which requires jointly preserving speech content, speaker identity, and temporal boundaries.

%

\definecolor{color2}{RGB}{098,190,166}
\definecolor{color3}{RGB}{253,186,107}
\definecolor{color4}{RGB}{235,096,070}
\definecolor{color5}{RGB}{111,146,225}
\definecolor{color6}{RGB}{175,135,220}
\definecolor{color7}{RGB}{143,203,107}
\definecolor{coloroutlier}{RGB}{176,176,176}

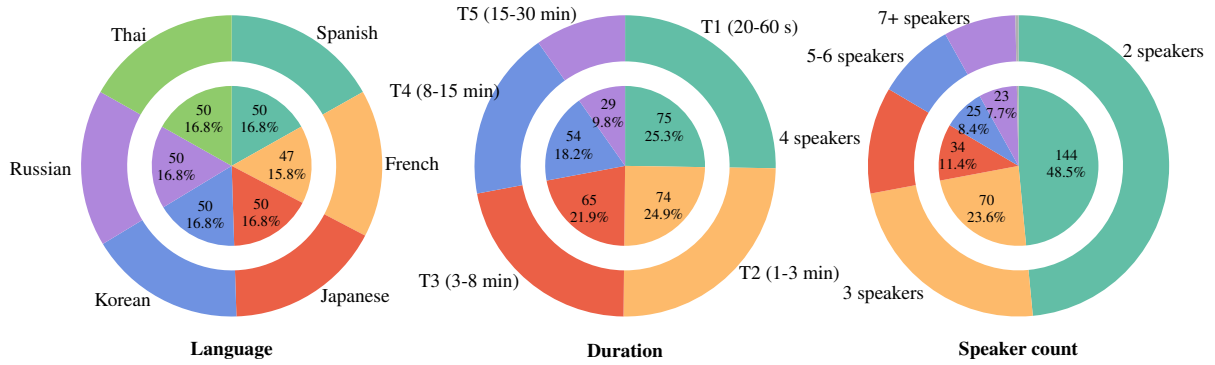
\begin{figure*}[t]
  \centering
  \resizebox{\textwidth}{!}{%
  \begin{tikzpicture}
  \begin{scope}[shift={(0.0,0)}]
    \fill[color2] (0,0) -- (90.000:1.3) arc[start angle=90.000, end angle=29.394, radius=1.3] -- cycle;
    \fill[color2] (90.000:2.45) arc[start angle=90.000, end angle=29.394, radius=2.45] -- (29.394:1.7) arc[start angle=29.394, end angle=90.000, radius=1.7] -- cycle;
    \fill[color3] (0,0) -- (29.394:1.3) arc[start angle=29.394, end angle=-27.576, radius=1.3] -- cycle;
    \fill[color3] (29.394:2.45) arc[start angle=29.394, end angle=-27.576, radius=2.45] -- (-27.576:1.7) arc[start angle=-27.576, end angle=29.394, radius=1.7] -- cycle;
    \fill[color4] (0,0) -- (-27.576:1.3) arc[start angle=-27.576, end angle=-88.182, radius=1.3] -- cycle;
    \fill[color4] (-27.576:2.45) arc[start angle=-27.576, end angle=-88.182, radius=2.45] -- (-88.182:1.7) arc[start angle=-88.182, end angle=-27.576, radius=1.7] -- cycle;
    \fill[color5] (0,0) -- (-88.182:1.3) arc[start angle=-88.182, end angle=-148.788, radius=1.3] -- cycle;
    \fill[color5] (-88.182:2.45) arc[start angle=-88.182, end angle=-148.788, radius=2.45] -- (-148.788:1.7) arc[start angle=-148.788, end angle=-88.182, radius=1.7] -- cycle;
    \fill[color6] (0,0) -- (-148.788:1.3) arc[start angle=-148.788, end angle=-209.394, radius=1.3] -- cycle;
    \fill[color6] (-148.788:2.45) arc[start angle=-148.788, end angle=-209.394, radius=2.45] -- (-209.394:1.7) arc[start angle=-209.394, end angle=-148.788, radius=1.7] -- cycle;
    \fill[color7] (0,0) -- (-209.394:1.3) arc[start angle=-209.394, end angle=-270.000, radius=1.3] -- cycle;
    \fill[color7] (-209.394:2.45) arc[start angle=-209.394, end angle=-270.000, radius=2.45] -- (-270.000:1.7) arc[start angle=-270.000, end angle=-209.394, radius=1.7] -- cycle;
    \node[align=center, font=\scriptsize] at (59.697:0.9) {50\\16.8\%};
    \node[align=center, font=\scriptsize] at (0.909:0.9) {47\\15.8\%};
    \node[align=center, font=\scriptsize] at (-57.879:0.9) {50\\16.8\%};
    \node[align=center, font=\scriptsize] at (-118.485:0.9) {50\\16.8\%};
    \node[align=center, font=\scriptsize] at (-179.091:0.9) {50\\16.8\%};
    \node[align=center, font=\scriptsize] at (-239.697:0.9) {50\\16.8\%};
    \node[anchor=west, font=\small] at (1.246,2.133) {Spanish};
    \node[anchor=west, font=\small] at (2.36,0.039) {French};
    \node[anchor=west, font=\small] at (1.313,-2.15) {Japanese};
    \node[anchor=east, font=\small] at (-1.178,-2.171) {Korean};
    \node[anchor=east, font=\small] at (-2.470,-0.039) {Russian};
    \node[anchor=east, font=\small] at (-1.246,2.133) {Thai};
    \node[font=\small] at (0,-3.000) {\textbf{Language}};
  \end{scope}
  \begin{scope}[shift={(6.4,0)}]
    \fill[color2] (0,0) -- (90.000:1.3) arc[start angle=90.000, end angle=-0.909, radius=1.3] -- cycle;
    \fill[color2] (90.000:2.45) arc[start angle=90.000, end angle=-0.909, radius=2.45] -- (-0.909:1.7) arc[start angle=-0.909, end angle=90.000, radius=1.7] -- cycle;
    \fill[color3] (0,0) -- (-0.909:1.3) arc[start angle=-0.909, end angle=-90.606, radius=1.3] -- cycle;
    \fill[color3] (-0.909:2.45) arc[start angle=-0.909, end angle=-90.606, radius=2.45] -- (-90.606:1.7) arc[start angle=-90.606, end angle=-0.909, radius=1.7] -- cycle;
    \fill[color4] (0,0) -- (-90.606:1.3) arc[start angle=-90.606, end angle=-169.394, radius=1.3] -- cycle;
    \fill[color4] (-90.606:2.45) arc[start angle=-90.606, end angle=-169.394, radius=2.45] -- (-169.394:1.7) arc[start angle=-169.394, end angle=-90.606, radius=1.7] -- cycle;
    \fill[color5] (0,0) -- (-169.394:1.3) arc[start angle=-169.394, end angle=-234.848, radius=1.3] -- cycle;
    \fill[color5] (-169.394:2.45) arc[start angle=-169.394, end angle=-234.848, radius=2.45] -- (-234.848:1.7) arc[start angle=-234.848, end angle=-169.394, radius=1.7] -- cycle;
    \fill[color6] (0,0) -- (-234.848:1.3) arc[start angle=-234.848, end angle=-270.000, radius=1.3] -- cycle;
    \fill[color6] (-234.848:2.45) arc[start angle=-234.848, end angle=-270.000, radius=2.45] -- (-270.000:1.7) arc[start angle=-270.000, end angle=-234.848, radius=1.7] -- cycle;
    \node[align=center, font=\scriptsize] at (44.545:0.9) {75\\25.3\%};
    \node[align=center, font=\scriptsize] at (-45.758:0.9) {74\\24.9\%};
    \node[align=center, font=\scriptsize] at (-130.000:0.9) {65\\21.9\%};
    \node[align=center, font=\scriptsize] at (-202.121:0.9) {54\\18.2\%};
    \node[align=center, font=\scriptsize] at (-252.424:0.9) {29\\9.8\%};
    \node[anchor=west, font=\small] at (1.10,2.233) {T1 (20-60 s)};
    \node[anchor=west, font=\small] at (1.723,-1.769) {T2 (1-3 min)};
    \node[anchor=east, font=\small] at (-1.588,-1.892) {T3 (3-8 min)};
    \node[anchor=east, font=\small] at (-1.9,1.2) {T4 (8-15 min)};
    \node[anchor=east, font=\small] at (-0.646,2.45) {T5 (15-30 min)};
    \node[font=\small] at (0,-3.000) {\textbf{Duration}};
  \end{scope}
  \begin{scope}[shift={(12.8,0)}]
    \fill[color2] (0,0) -- (90.000:1.3) arc[start angle=90.000, end angle=-84.545, radius=1.3] -- cycle;
    \fill[color2] (90.000:2.45) arc[start angle=90.000, end angle=-84.545, radius=2.45] -- (-84.545:1.7) arc[start angle=-84.545, end angle=90.000, radius=1.7] -- cycle;
    \fill[color3] (0,0) -- (-84.545:1.3) arc[start angle=-84.545, end angle=-169.394, radius=1.3] -- cycle;
    \fill[color3] (-84.545:2.45) arc[start angle=-84.545, end angle=-169.394, radius=2.45] -- (-169.394:1.7) arc[start angle=-169.394, end angle=-84.545, radius=1.7] -- cycle;
    \fill[color4] (0,0) -- (-169.394:1.3) arc[start angle=-169.394, end angle=-210.606, radius=1.3] -- cycle;
    \fill[color4] (-169.394:2.45) arc[start angle=-169.394, end angle=-210.606, radius=2.45] -- (-210.606:1.7) arc[start angle=-210.606, end angle=-169.394, radius=1.7] -- cycle;
    \fill[color5] (0,0) -- (-210.606:1.3) arc[start angle=-210.606, end angle=-240.909, radius=1.3] -- cycle;
    \fill[color5] (-210.606:2.45) arc[start angle=-210.606, end angle=-240.909, radius=2.45] -- (-240.909:1.7) arc[start angle=-240.909, end angle=-210.606, radius=1.7] -- cycle;
    \fill[color6] (0,0) -- (-240.909:1.3) arc[start angle=-240.909, end angle=-268.788, radius=1.3] -- cycle;
    \fill[color6] (-240.909:2.45) arc[start angle=-240.909, end angle=-268.788, radius=2.45] -- (-268.788:1.7) arc[start angle=-268.788, end angle=-240.909, radius=1.7] -- cycle;
    \fill[coloroutlier] (0,0) -- (-268.788:1.3) arc[start angle=-268.788, end angle=-270.000, radius=1.3] -- cycle;
    \fill[coloroutlier] (-268.788:2.45) arc[start angle=-268.788, end angle=-270.000, radius=2.45] -- (-270.000:1.7) arc[start angle=-270.000, end angle=-268.788, radius=1.7] -- cycle;
    \node[align=center, font=\scriptsize] at (2.727:0.78) {144\\48.5\%};
    \node[align=center, font=\scriptsize] at (-126.970:0.86) {70\\23.6\%};
    \node[align=center, font=\scriptsize] at (-190.000:1.0) {34\\11.4\%};
    \node[align=center, font=\scriptsize] at (-225.758:1.040) {25\\8.4\%};
    \node[align=center, font=\scriptsize] at (-254.848:1.040) {23\\7.7\%};
    \node[anchor=west, font=\small] at (1.56,1.85) {2 speakers};
    \node[anchor=east, font=\small] at (-1.4,-2.1) {3 speakers};
    \node[anchor=east, font=\small] at (-2.432,0.429) {4 speakers};
    \node[anchor=east, font=\small] at (-1.723,1.769) {5-6 speakers};
    \node[anchor=east, font=\small] at (-0.646,2.384) {7+ speakers};
    \node[font=\small] at (0,-3.000) {\textbf{Speaker count}};
  \end{scope}
  \end{tikzpicture}%
  }
  \caption{Benchmark composition of \name in three views: languages, audio duration and speaker count. For each view we shows case quantity, category labels, and necessary explanations. }
  \label{fig:category}
\end{figure*}

\subsection{Multilingual Multi-Speaker ASR}

Multi-speaker ASR aims to recover speaker attributed transcripts from conversations containing multiple, potentially overlapping speakers. Unlike conventional ASR, which determines what was said, or speaker diarization, which determines who spoke when, multi-speaker ASR must jointly resolve speech content and speaker attribution.

Meeting corpora and challenges such as AMI \citep{carletta2005ami}, CHiME-6 \citep{watanabe2020chime6}, Aishell-3 \citep{shi2021aishell}, AliMeeting \citep{yu2022m2met}, and NOTSOFAR-1 \citep{vinnikov2024notsofar} have established standard evaluation settings for multi-speaker transcription. 
Recent efforts such as Muscat \citep{sinhamahapatra2026muscat} and MLC-SLM \citep{mu2026summary} further extend conversational speech modeling to multilingual conditions. Nevertheless, existing benchmarks remain concentrated on short-form audio, acoustic domains, or specialized sepeech systems without omni LLMs.

Multi speaker ASR methods have progressed from modular diarization, separation, and ASR pipelines to permutation-invariant and serialized end-to-end modeling \citep{yu2017pit, kanda2020sot}, followed by LLM-based systems such as MT-LLM and SpeakerLM \citep{meng2025mtllm, yin2026speakerlm}. Although these systems demonstrate the feasibility of LLM-based multi-speaker transcription, existing evaluations provide only limited evidence about general-purpose omni LLMs \citep{kwon2026m3slu, tawara2026whospoke}. Their robustness across languages, recording durations, speaker counts, and overlap conditions therefore remains insufficiently characterized.

\section{\textsc{OmniLingual-MsASR}}

\subsection{Overview}

In this section, we introduce \name, a multimodal, multilingual, and long-form MsASR benchmark designed to probe the capability boundaries of omni LLMs in real-world audio-visual transcription. As shown in Fig. \ref{fig:category}, \name comprises 297 instances spanning six diverse languages, with 50 instances each for Spanish, Japanese, Korean, Thai, and Russian, alongside 47 instances for French.

To assess model scalability across extended temporal contexts, instance durations range from 20 s short clips to 30 min long-form videos. Considering the sample distribution across durations, we stratify \name into five tiers partitioned by cutoffs at 1, 3, 8, and 15 minutes, as shown in the middle panel of Fig. \ref{fig:category}. 

Furthermore, speaker density serves as a critical indicator of MsASR complexity. \name scales up to 20 speakers per instance to challenge omni LLMs at their limits. To maintain balanced sample sizes across subsets, we aggregate instances into discrete categories, specifically consolidating cases with five to six speakers and those with seven or more speakers, thereby facilitating a robust evaluation of speaker identity, detailed in the right panel of Fig. \ref{fig:category}.

Notably, \name provides parallel evaluation for both audio-only and audio-visual MsASR. For human listeners, speaker identification relies on purely acoustic signals is inherently challenging, especially under high vocal similarity. Conversely, visual information such as facial appearance provides crucial speaker cues. 
Through this dual-eval setup, we explore whether omni LLMs have surpassed the inflection point where visual context effectively enhances speaker identification, or redundant visual signals conversely introduce noise that impact model performance.

\subsection{Data Source}
To ensure public accessibility and reproducibility, we collected long-form video candidates ranging from 1.5 to 2 hours from YouTube, initially sourcing 10 videos per language. We then conducted manual filtering to discard videos with fewer than two speakers, as well as variety shows with excessive speech overlap, which poses significant annotation challenges. The resulting corpus consists of multilingual, multi-speaker long-form videos spanning diverse domains, including meetings, podcasts, news broadcasts, TV dramas, speeches, prayers, and daily vlogs.

\begin{figure}[t]
  \centering
  \includegraphics[width=\columnwidth]{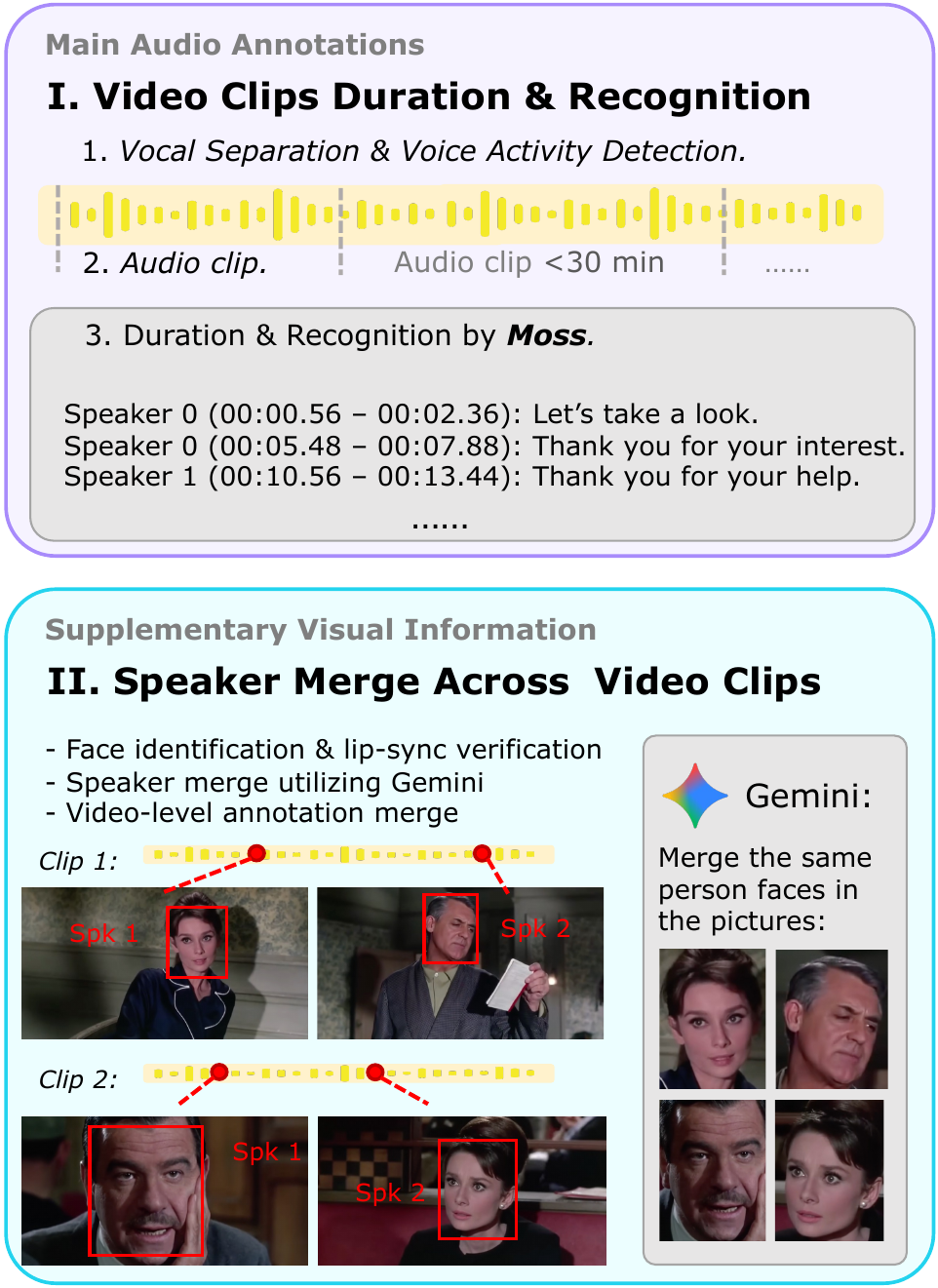}
  \caption{Overview of automatic data pipeline. The pipeline first leverages audio models to segment and annotate individual clips, followed by visual models to identify and consolidate cross-clip speaker identities into pseudo MsASR annotations of long-form videos.}
  \label{fig:pipeline}
\end{figure}

\subsection{Automatic Data Pipeline}

To generate high-quality pseudo labels, our automated pipeline leverages frontier models for scalable annotation, as illustrated in Fig. \ref{fig:pipeline}. To annotate long-form videos that exceed model context limits, the pipeline first acoustically segments raw videos into clips, subsequently generates clip-level annotations using \texttt{MOSS Transcribe Diarize 0.9B}, and finally reconciles cross-clip speaker identities using complementary visual cues to form unified long-form annotations of the whole videos.

\paragraph{Audio-Centric Pseudo Labeling.}
Given the diverse acoustic noise in YouTube videos, we first apply \texttt{MossFormer2\_SE} for speech enhancement to improve annotation robustness. 
Because full-length recordings ($\sim$2 hours) exceed the context window of \texttt{MOSS Transcribe Diarize 0.9B}, we segment the video streams into parallel video and audio sub-30-minute clips. To prevent bisecting utterances, we utlize Voice Activity Detection and split clips exclusively along silence intervals. Finally, each clip is processed by \texttt{MOSS Transcribe Diarize 0.9B} to extract multi-speaker ASR pseudo-labels, comprising speaker durations and transcriptions.

\paragraph{Visual-Aided Cross-Clip Identity Merging.}
Independently annotating individual clips introduces fragmented, local speaker IDs for the same individual across clips as shown in Fig. \ref{fig:pipeline}. Audio-only clustering struggles as fixed acoustic embedding thresholds fail to balance inner-speaker diversity against inter-speaker timbre similarity. 
Therefore, we leverage visual facial cues to globally align speaker identities. Specifically, we utilize SyncNet \citep{Chung16a} for lip-sync verification to identify active speaking faces, and then prompt \texttt{Gemini 3.1 Pro} with paired local speaker IDs and face crops to resolve cross-clip identities. 
Then, utterance timestamps are subsequently shifted to the global timeline. To ensure balanced durations and category distributions, benchmark samples are sampled as sub-segments with potential overlap from these merged video annotations.
 
\subsection{Linguistic Expert Annotation}

Following the automatic data pipeline, we implemented a rigorous three-stage verification protocol: Annotation, Review, and Spot-checking. 

The first two stages were conducted by linguistic experts, who verified the alignment of speaker identities, timestamps, and transcriptions against the corresponding videos. Specifically, annotators first corrected the pipeline outputs, after which an independent, disjoint pool of experts reviewed the annotations to ensure quality. 
In the final stage, the authors audited a randomly sampled subset. Constrained by multilingual proficiency, this check primarily focused on language-agnostic attributes, namely speaker labels and timestamps. Detailed annotation guidelines are provided in Appendix \ref{append:guideline}.

\subsection{Ethical and Privacy Audit}
Following expert verification, we collected approximately 60 annotated instances per language, from which 50 instances were carefully curated to form a preliminary benchmark, prioritizing balanced temporal distribution and content diversity. 

Subsequently, we conducted an ethical and privacy audit. During this stage, three French instances were found to contain Personally Identifiable Information (PII), specifically authentic social media handles and email addresses. To mitigate privacy leakage risks and adhere to ethical research standards, we removed these instances, yielding a finalized benchmark of 297 high quality samples.

\section{Experiments}

\subsection{Experimental Setup}

\paragraph{Baseline models.}
We conduct extensive evaluations of frontier omni-modal LLMs.
Open-source models include task-specific \texttt{MOSS Transcribe Diarize 0.9B} \citep{ai2026moss} and omni LLM \texttt{Qwen3-Omni-30B-A3B} \citep{xu2025qwen3}. 
We also evaluate top-tier commercial models including \texttt{Gemini 3.1 Pro}, \texttt{Gemini 3.7 Flash}, \texttt{Gemini 3.8 Flash}, \texttt{Qwen3.5-Omni-Plus} \citep{team2026qwen3}, \texttt{Qwen3.8-Omni-Flash} \citep{team2026qwen38}, \texttt{Muse Spark 1.2}, and \texttt{Doubao-Seed-2.0-Lite}. 

\paragraph{Inference hyperparameter.} For all models, we retain the default inference hyperparameter settings to elicit optimal model performance.
The open-source model runs on 8x A100 GPUs.

\subsection{Evaluate Metrics}
To evaluate the MsASR performance, we use four metrics: parse rate, diarization error rate (DER) \citep{shafey2019joint}, concatenated minimum-permutation word error rate (cpWER) \citep{watanabe2020chime}, and time-constrained permutation WER (tcpWER) \citep{von2023meeteval}. 

\paragraph{Parse rate} measures the instruction following capability of omni LLMs. We require omni LLMs to output speaker id, timestamp, and text as follow:
\begin{tcolorbox}[breakable, colback=gray!5, colframe=gray!50, title=Output format, left=1.5mm]
\small
\begin{Verbatim}
<soc>
    <sos>
         <start_time>text<end_time><speakerX>        
    <eos>
    <sos>
         <start_time>text<end_time><speakerX>
    <eos>
    ...
<eoc>
\end{Verbatim}
\label{tab:format}
\end{tcolorbox}
Few cases that models fail to follow the instruction or degenerate into repetitive generation, rendering their outputs unparsable. We exclude such cases from the three WER metrics, so that a few outliers with extremely high error rates do not distort the overall assessment of model capability.

\paragraph{DER} measures who spoke when while ignoring the transcribed content, which is defined as:
\begin{equation}
    \text{DER}=\frac{T_{\text{miss}}+T_{\text{fa}}+T_{\text{conf}}}{T_{\text{total}}}
    \label{ep:der}
\end{equation}
where $T_{\text{miss}}$ is the missed-speech time, $T_{\text{fa}}$ is the false-alarm time, and $T_{\text{conf}}$ is the speaker-confusion time, while $T_{\text{total}}$ denotes the total reference speaking time.
We use a forgiveness collar of $0.5$ s around each reference boundary in DER.

\paragraph{cpWER} jointly scores transcription and speaker attribution while ignoring the timestamps:
\begin{equation}
    \text{cpWER}=\frac{S+D+I}{N}
\end{equation}
where $S$, $D$, and $I$ are the total numbers of substituted, deleted, and inserted words summed over all speakers under the speaker pairing that minimizes the total edit distance between the per-speaker concatenated transcripts, and $N$ is the total number of reference words.

\paragraph{tcpWER} augments cpWER by allowing a reference and hypothesis word to be aligned only when their timestamps are close, so temporally misplaced words are penalized even if the text is correct:
\begin{equation}
    \text{cpWER}=\frac{S_\tau+D_\tau+I_\tau}{N}
\end{equation}
where $S_\tau$, $D_\tau$, and $I_\tau$ are the substituted, deleted, and inserted word counts obtained when only words whose timestamps lie within the collar $\tau$ are allowed to be aligned, and $N$ is the total number of reference words as above. We use a temporal collar of $\tau = 5$ s in tcpWER.

\begin{table*}[t]
\centering
\setlength{\tabcolsep}{4pt}
\renewcommand{\arraystretch}{1.15}
\resizebox{\textwidth}{!}{%
\begin{tabular}{l cccc cccc cccc}
\toprule
\multirow{2}{*}{\textbf{Model}}
 & \multicolumn{4}{c}{\textbf{es}} & \multicolumn{4}{c}{\textbf{fr}} & \multicolumn{4}{c}{\textbf{ja}} \\
\cmidrule(lr){2-5}\cmidrule(lr){6-9}\cmidrule(lr){10-13}
 & Parse & DER & cpWER & tcpWER & Parse & DER & cpWER & tcpWER & Parse & DER & cpWER & tcpWER \\
\midrule
\multicolumn{13}{l}{\cellcolor{black!10} \textbf{\textit{I. Frontier Commercial Omni-Modal Large Language Models}}} \\
gemini 3.1 pro & 90 & 0.628 & \underline{0.257} & 1.385 & 88 & 0.567 & 0.373 & 1.248 & 90 & 0.599 & 0.423 & 1.311 \\
gemini 3.7 flash & 84 & 0.516 & 0.294 & 0.921 & 76 & 0.637 & \textbf{0.348} & 1.050 & 88 & 0.565 & \underline{0.410} & 0.987 \\
gemini 3.8 flash & \underline{98} & 0.590 & \textbf{0.230} & 1.079 & \underline{98} & 0.738 & 0.440 & 1.322 & 96 & 0.677 & 0.441 & 1.325 \\
qwen3.5-omni-plus & 76 & \underline{0.383} & 0.395 & \underline{0.864} & 84 & \underline{0.360} & 0.439 & \underline{0.799} & 94 & 0.464 & 0.577 & \underline{0.873} \\
qwen3.8-omni-flash & \textbf{100} & \textbf{0.160} & 0.267 & \textbf{0.344} & \textbf{100} & \textbf{0.178} & \underline{0.362} & \textbf{0.453} & \underline{98} & \textbf{0.085} & \textbf{0.329} & \textbf{0.360} \\
muse spark 1.2 & \textbf{100} & 0.696 & 0.678 & 1.038 & \underline{98} & 0.779 & 0.782 & 1.090 & \textbf{100} & 0.757 & 0.804 & 1.034 \\
doubao-seed-2.0-lite & 86 & 0.390 & 0.288 & 0.865 & 90 & 0.532 & 0.437 & 1.041 & 90 & \underline{0.447} & 0.424 & 0.990 \\
\multicolumn{13}{l}{\cellcolor{black!10} \textbf{\textit{II. Open Source Omni-Modal Large Language Model}}} \\
qwen3-omni-30b-thinking & 80 & 0.532 & 0.469 & 0.996 & 77 & 0.694 & 0.672 & 1.251 & 74 & 0.561 & 0.597 & 1.058 \\
\multicolumn{13}{l}{\cellcolor{black!10} \textbf{\textit{III. Expert Diarization Model}}} \\
moss-transcribe-diarize & 100 & 0.115 & 0.206 & 0.213 & 100 & 0.187 & 0.348 & 0.377 & 100 & 0.059 & 0.273 & 0.276 \\
\midrule
\multirow{2}{*}{\textbf{Model}}
& \multicolumn{4}{c}{\textbf{ko}} & \multicolumn{4}{c}{\textbf{ru}} & \multicolumn{4}{c}{\textbf{th}} \\
\cmidrule(lr){2-5}\cmidrule(lr){6-9}\cmidrule(lr){10-13}
 & Parse & DER & cpWER & tcpWER & Parse & DER & cpWER & tcpWER & Parse & DER & cpWER & tcpWER \\
\midrule
\multicolumn{13}{l}{\cellcolor{black!10} \textbf{\textit{I. Frontier Commercial Omni-Modal Large Language Models}}} \\
gemini 3.1 pro & 92 & 0.589 & \underline{0.302} & 1.275 & \underline{90} & 0.567 & 0.436 & 1.344 & 90 & 0.806 & 1.352 & 1.828 \\
gemini 3.7 flash & 70 & 0.662 & 0.315 & 1.044 & 86 & 0.737 & 0.453 & 1.230 & 88 & 0.684 & 1.654 & 2.099 \\
gemini 3.8 flash & 94 & 0.697 & 0.317 & 1.185 & \textbf{100} & 0.662 & 0.402 & 1.189 & \underline{96} & 1.059 & 2.457 & 2.988 \\
qwen3.5-omni-plus & \underline{96} & 0.634 & 0.611 & \underline{0.887} & \underline{90} & \underline{0.305} & \underline{0.388} & \underline{0.676} & 90 & 0.855 & 1.217 & 1.432 \\
qwen3.8-omni-flash & \textbf{100} & \textbf{0.171} & \textbf{0.277} & \textbf{0.405} & \textbf{100} & \textbf{0.185} & \textbf{0.305} & \textbf{0.491} & \textbf{98} & \underline{0.682} & 1.323 & 1.412 \\
muse spark 1.2 & \underline{96} & 0.763 & 0.807 & 1.066 & \textbf{100} & 0.722 & 0.780 & 1.040 & \textbf{98} & 0.732 & \textbf{0.995} & \textbf{1.117} \\
doubao-seed-2.0-lite & 88 & \underline{0.517} & 0.467 & 0.952 & \underline{90} & 0.575 & 0.466 & 1.042 & 88 & \textbf{0.619} & \underline{1.011} & \underline{1.306} \\
\multicolumn{13}{l}{\cellcolor{black!10} \textbf{\textit{II. Open Source Omni-Modal Large Language Model}}} \\
qwen3-omni-30b-thinking & 82 & 0.826 & 0.673 & 1.280 & 82 & 0.661 & 0.769 & 1.452 & 58 & 0.804 & 1.081 & 1.230 \\
\multicolumn{13}{l}{\cellcolor{black!10} \textbf{\textit{III. Expert Diarization Model}}} \\
moss-transcribe-diarize & 96 & 0.091 & 0.197 & 0.202 & 100 & 0.163 & 0.271 & 0.298 & 100 & 0.420 & 0.841 & 0.929 \\
\bottomrule
\end{tabular}%
}
\caption{Multi-speaker ASR results of six languages: Spanish (es), French (fr), Japanese (ja), Korean (ko), Russian (ru), and Thai (th).
Parse(\%) is the format parse rate of model output. Lower is better for three multi-speaker speech recognition metrics: DER($\downarrow$), cpWER($\downarrow$), and tcpWER($\downarrow$).}
\label{tab:msasr-lang-view}
\end{table*}

\section{Results and Deep Analysis}

\subsection{Main Results}
In this section, we present the main evaluation results of frontier omni-LLMs across six languages on \name.
First, output parsing rates reveal \textit{disparities in multilingual instruction following}. While most commercial models achieve consistently high parse rates, notable degradations occur with \texttt{Gemini 3.7 Flash} on fr and ko, as well as \texttt{Qwen3.5-Omni-Plus} on es, both falling below 80\%. This highlights the uneven multilingual instruction-following capability in omni LLMs.

Second, current models \textit{excel at transcription but struggle with temporal alignment}. Across es, fr, ja, ko, and ru, cpWER is consistently and substantially lower than both DER and tcpWER for the Gemini, Muse Spark, and Doubao series. This indicates that while these models can differentiate speakers and transcribe linguistic content, they lack the capability to precisely estimate time boundaries. In contrast, the Qwen series significantly outperforms other models on DER, demonstrating a clear advantage in speech duration.

Third, \textit{Thai presents unique acoustic challenges}. Performance on Thai is markedly inferior across all evaluated models. We attribute this to intrinsic video difficulty: human inspection reveals higher timbre similarity among Thai speakers, which considerably complicates speaker identification.

Fourth, \textit{frontier commercial omni-LLMs substantially outperform their open-source counterparts and approach specialized diarization models.} 
Notably, \texttt{Qwen3-Omni-30B} falls markedly behind the best commercial results. 
Furthermore, while general-purpose omni-LLMs are comparable to the task-specific \texttt{MOSS-Transcribe-Diarize-0.9B}, MOSS still retains a clear lead over nearly all competing models in all languages.

Overall, \texttt{MOSS-Transcribe-Diarize-7B} and \texttt{Qwen3.8-Omni-Flash} delivers the strongest overall performance on \textsc{OmniLingual-MsASR} benchmark, consistently ranking in the top three across virtually all evaluation metrics.

\definecolor{color2}{RGB}{098,190,166}
\definecolor{color3}{RGB}{253,186,107}
\definecolor{color4}{RGB}{235,096,070}
\definecolor{color5}{RGB}{111,146,225}
\definecolor{color6}{RGB}{175,135,220}
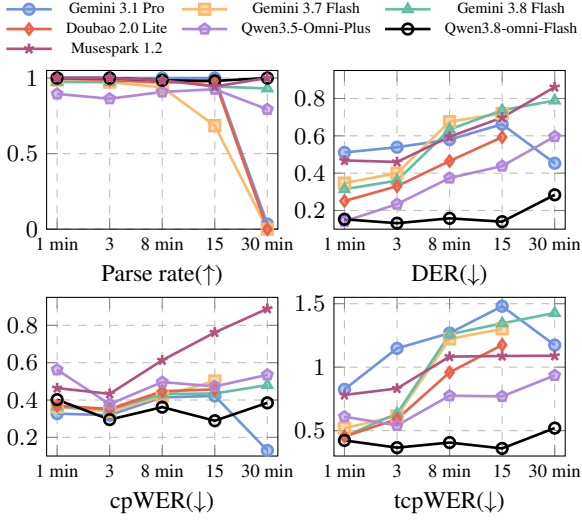
\begin{figure}[t]
\begin{tikzpicture}
\footnotesize{
\begin{axis}[
at={(0cm,0cm)},
    ymajorgrids,
    xmajorgrids,
    grid style=dashed,
    width=0.29\textwidth, height=0.23\textwidth,
    legend columns=3,
    legend entries={Gemini 3.1 Pro,Gemini 3.7 Flash,Gemini 3.8 Flash,Doubao 2.0 Lite,Qwen3.5-Omni-Plus,Qwen3.8-omni-Flash,Musespark 1.2},
    legend style={fill opacity=0.5,text opacity=1,draw=none,line width=1pt,at={(-0.22,1.49)},anchor=north west,nodes={scale=0.65,transform shape},column sep=0.4em},
    xmin=0.8, xmax=5.2,
    ymin=0, ymax=1.05,
    xtick={1,2,3,4,5},
    xticklabels={\scriptsize{1 min},\scriptsize{3},\scriptsize{8 min},\scriptsize{15},\scriptsize{30 min}},
    xlabel=\footnotesize{Parse rate($\uparrow$)},
    xlabel style={yshift=0.1em,align=center},
    scaled ticks=false,
    every axis plot/.append style={mark options={fill opacity=0.5,draw opacity=1}},
    ]
    \addplot[color5,mark=*,line width=1pt] coordinates {(1,1.000) (2,1.000) (3,1.000) (4,1.000) (5,0.034)};
    \addplot[color3,mark=square*,line width=1pt] coordinates {(1,0.987) (2,0.973) (3,0.938) (4,0.685) (5,0.000)};
    \addplot[color2,mark=triangle*,line width=1pt] coordinates {(1,0.974) (2,0.973) (3,1.000) (4,0.944) (5,0.931)};
    \addplot[color4,mark=diamond*,line width=1pt] coordinates {(1,1.000) (2,0.986) (3,0.969) (4,0.981) (5,0.000)};
    \addplot[color6,mark=pentagon*,line width=1pt] coordinates {(1,0.895) (2,0.863) (3,0.908) (4,0.926) (5,0.793)};
    \addplot[black,mark=o,line width=1pt] coordinates {(1,1.000) (2,1.000) (3,0.985) (4,0.981) (5,1.000)};
    \addplot[magenta!70!black,mark=star,line width=1pt] coordinates {(1,1.000) (2,1.000) (3,0.985) (4,0.944) (5,1.000)};
\end{axis}
}
\footnotesize{
\begin{axis}[
at={(3.8cm,0cm)},
    ymajorgrids,
    xmajorgrids,
    grid style=dashed,
    width=0.29\textwidth, height=0.23\textwidth,
    legend columns=-1,
    xmin=0.8, xmax=5.2,
    ymin=0.1, ymax=0.95,
    xtick={1,2,3,4,5},
    xticklabels={\scriptsize{1 min},\scriptsize{3},\scriptsize{8 min},\scriptsize{15},\scriptsize{30 min}},
    xlabel=\footnotesize{DER($\downarrow$)},
    xlabel style={yshift=0.1em,align=center},
    scaled ticks=false,
    every axis plot/.append style={mark options={fill opacity=0.5,draw opacity=1}},
    ]
    \addplot[color5,mark=*,line width=1pt] coordinates {(1,0.511) (2,0.539) (3,0.580) (4,0.662) (5,0.453)};
    \addplot[color3,mark=square*,line width=1pt] coordinates {(1,0.347) (2,0.400) (3,0.676) (4,0.720)};
    \addplot[color2,mark=triangle*,line width=1pt] coordinates {(1,0.314) (2,0.360) (3,0.634) (4,0.737) (5,0.789)};
    \addplot[color4,mark=diamond*,line width=1pt] coordinates {(1,0.251) (2,0.330) (3,0.465) (4,0.593)};
    \addplot[color6,mark=pentagon*,line width=1pt] coordinates {(1,0.142) (2,0.234) (3,0.374) (4,0.438) (5,0.596)};
    \addplot[black,mark=o,line width=1pt] coordinates {(1,0.153) (2,0.132) (3,0.158) (4,0.140) (5,0.284)};
    \addplot[magenta!70!black,mark=star,line width=1pt] coordinates {(1,0.468) (2,0.460) (3,0.595) (4,0.698) (5,0.861)};
\end{axis}
}

\footnotesize{
\begin{axis}[
at={(0cm,-3.0cm)},
    ymajorgrids,
    xmajorgrids,
    grid style=dashed,
    width=0.29\textwidth, height=0.23\textwidth,
    legend columns=-1,
    xmin=0.8, xmax=5.2,
    ymin=0.1, ymax=0.95,
    xtick={1,2,3,4,5},
    xticklabels={\scriptsize{1 min},\scriptsize{3},\scriptsize{8 min},\scriptsize{15},\scriptsize{30 min}},
    xlabel=\footnotesize{cpWER($\downarrow$)},
    xlabel style={yshift=0.1em,align=center},
    scaled ticks=false,
    every axis plot/.append style={mark options={fill opacity=0.5,draw opacity=1}},
    ]
    \addplot[color5,mark=*,line width=1pt] coordinates {(1,0.327) (2,0.319) (3,0.414) (4,0.422) (5,0.130)};
    \addplot[color3,mark=square*,line width=1pt] coordinates {(1,0.362) (2,0.338) (3,0.421) (4,0.502)};
    \addplot[color2,mark=triangle*,line width=1pt] coordinates {(1,0.365) (2,0.349) (3,0.432) (4,0.435) (5,0.480)};
    \addplot[color4,mark=diamond*,line width=1pt] coordinates {(1,0.370) (2,0.352) (3,0.447) (4,0.457)};
    \addplot[color6,mark=pentagon*,line width=1pt] coordinates {(1,0.562) (2,0.378) (3,0.496) (4,0.473) (5,0.535)};
    \addplot[black,mark=o,line width=1pt] coordinates {(1,0.402) (2,0.295) (3,0.362) (4,0.289) (5,0.385)};
    \addplot[magenta!70!black,mark=star,line width=1pt] coordinates {(1,0.464) (2,0.433) (3,0.613) (4,0.762) (5,0.889)};
\end{axis}
}
\footnotesize{
\begin{axis}[
at={(3.8cm,-3.0cm)},
    ymajorgrids,
    xmajorgrids,
    grid style=dashed,
    width=0.29\textwidth, height=0.23\textwidth,
    legend columns=-1,
    xmin=0.8, xmax=5.2,
    ymin=0.3, ymax=1.55,
    xtick={1,2,3,4,5},
    xticklabels={\scriptsize{1 min},\scriptsize{3},\scriptsize{8 min},\scriptsize{15},\scriptsize{30 min}},
    xlabel=\footnotesize{tcpWER($\downarrow$)},
    xlabel style={yshift=0.1em,align=center},
    scaled ticks=false,
    every axis plot/.append style={mark options={fill opacity=0.5,draw opacity=1}},
    ]
    \addplot[color5,mark=*,line width=1pt] coordinates {(1,0.825) (2,1.148) (3,1.268) (4,1.480) (5,1.174)};
    \addplot[color3,mark=square*,line width=1pt] coordinates {(1,0.519) (2,0.616) (3,1.222) (4,1.302)};
    \addplot[color2,mark=triangle*,line width=1pt] coordinates {(1,0.454) (2,0.635) (3,1.259) (4,1.345) (5,1.427)};
    \addplot[color4,mark=diamond*,line width=1pt] coordinates {(1,0.453) (2,0.591) (3,0.959) (4,1.175)};
    \addplot[color6,mark=pentagon*,line width=1pt] coordinates {(1,0.610) (2,0.541) (3,0.775) (4,0.770) (5,0.935)};
    \addplot[black,mark=o,line width=1pt] coordinates {(1,0.423) (2,0.366) (3,0.406) (4,0.360) (5,0.520)};
    \addplot[magenta!70!black,mark=star,line width=1pt] coordinates {(1,0.781) (2,0.832) (3,1.083) (4,1.088) (5,1.090)};
\end{axis}
}
\end{tikzpicture}

\caption{Performance across audio durations on \name. Omni LLMs performance generally degrade as audio duration increase. Audio duration categories detalied at Fig. \ref{fig:category}.}
\label{fig:duration}
\end{figure}

\subsection{Analysis of Audio Duration}

\paragraph{Analysis of DER components.} While our primary results show that current models excel at transcription, they struggle with temporal alignment. To determine whether the high DER stems from speaker misidentification or imprecise timestamp prediction, we dissect DER into its three components as described in Eq. \ref{ep:der}: missed speech $T_{\text{miss}}$, false alarm $T_{\text{fa}}$, and speaker confusion $T_{\text{conf}}$. Averaged across seven models, missed speech constitutes the majority 31.7\% of the overall 54.4\% DER, whereas false alarm and speaker confusion account for only 10.2\% and 12.5\%, respectively. 
This further confirms that the \textit{bottleneck of current models lies primarily in accurate timestamp prediction} rather than speaker discrimination.

\paragraph{Model performance with longer audio duration.} Long-form audio understanding remains a key challenge for omni LLMs. In this section, we investigate their MsASR performance across varying audio lengths. 
Specifically, \textsc{OmniLingual-MsASR} spans audio durations from 20 seconds to 30 minutes, partitioned into five duration buckets, as detailed in middle sub-figure of Fig \ref{fig:category}.

As illustrated in Fig. \ref{fig:duration}, \textit{model performance generally degrades as audio duration increases}. Notably, \texttt{Doubao-Seed-2.0-Lite} and \texttt{Gemini 3.7 Flash} exhibit a marked drop in parseability rate within the 15–30 minute range. Furthermore, the time-sensitive metrics DER and tcpWER consistently deteriorate on longer inputs, whereas this trend is far less pronounced for cpWER. This disparity indicates that temporal perception errors in omni LLMs compound over extended contexts, leading to degraded timestamp accuracy.

\definecolor{color2}{RGB}{098,190,166}
\definecolor{color3}{RGB}{253,186,107}
\definecolor{color4}{RGB}{235,096,070}
\definecolor{color5}{RGB}{111,146,225}
\definecolor{color6}{RGB}{175,135,220}
\begin{figure}[t]
\begin{tikzpicture}
\footnotesize{
\begin{axis}[
at={(0cm,0cm)},
    ymajorgrids,
    xmajorgrids,
    grid style=dashed,
    width=0.29\textwidth, height=0.23\textwidth,
    legend columns=3,
    legend entries={Gemini 3.1 Pro,Gemini 3.7 Flash,Gemini 3.8 Flash,Doubao 2.0 Lite,Qwen3.5-Omni-Plus,Qwen3.8-omni-Flash,Musespark 1.2},
    legend style={fill opacity=0.5,text opacity=1,draw=none,line width=1pt,at={(-0.22,1.49)},anchor=north west,nodes={scale=0.65,transform shape},column sep=0.4em},
    xmin=0.8, xmax=5.2,
    ymin=0, ymax=1.05,
    xtick={1,2,3,4,5},
    xticklabels={\scriptsize{2},\scriptsize{3},\scriptsize{4},\scriptsize{5-6},\scriptsize{7+}},
    xlabel=\footnotesize{Parse rate($\uparrow$)},
    xlabel style={yshift=0.1em,align=center},
    scaled ticks=false,
    every axis plot/.append style={mark options={fill opacity=0.5,draw opacity=1}},
    ]
    \addplot[color5,mark=*,line width=1pt] coordinates {(1,0.938) (2,0.957) (3,0.882) (4,0.840) (5,0.696)};
    \addplot[color3,mark=square*,line width=1pt] coordinates {(1,0.875) (2,0.857) (3,0.794) (4,0.600) (5,0.696)};
    \addplot[color2,mark=triangle*,line width=1pt] coordinates {(1,0.965) (2,0.986) (3,1.000) (4,0.960) (5,0.913)};
    \addplot[color4,mark=diamond*,line width=1pt] coordinates {(1,0.917) (2,0.943) (3,0.853) (4,0.840) (5,0.696)};
    \addplot[color6,mark=pentagon*,line width=1pt] coordinates {(1,0.861) (2,0.871) (3,0.941) (4,0.960) (5,0.913)};
    \addplot[black,mark=o,line width=1pt] coordinates {(1,1.000) (2,0.986) (3,1.000) (4,1.000) (5,0.957)};
    \addplot[magenta!70!black,mark=star,line width=1pt] coordinates {(1,0.979) (2,0.986) (3,1.000) (4,1.000) (5,1.000)};
\end{axis}
}
\footnotesize{
\begin{axis}[
at={(3.8cm,0cm)},
    ymajorgrids,
    xmajorgrids,
    grid style=dashed,
    width=0.29\textwidth, height=0.23\textwidth,
    legend columns=-1,
    xmin=0.8, xmax=5.2,
    ymin=0.1, ymax=1.15,
    xtick={1,2,3,4,5},
    xticklabels={\scriptsize{2},\scriptsize{3},\scriptsize{4},\scriptsize{5-6},\scriptsize{7+}},
    xlabel=\footnotesize{DER($\downarrow$)},
    xlabel style={yshift=0.1em,align=center},
    scaled ticks=false,
    every axis plot/.append style={mark options={fill opacity=0.5,draw opacity=1}},
    ]
    \addplot[color5,mark=*,line width=1pt] coordinates {(1,0.588) (2,0.549) (3,0.717) (4,0.600) (5,1.009)};
    \addplot[color3,mark=square*,line width=1pt] coordinates {(1,0.559) (2,0.601) (3,0.627) (4,0.777) (5,1.084)};
    \addplot[color2,mark=triangle*,line width=1pt] coordinates {(1,0.591) (2,0.645) (3,0.782) (4,0.885) (5,0.976)};
    \addplot[color4,mark=diamond*,line width=1pt] coordinates {(1,0.419) (2,0.500) (3,0.608) (4,0.620) (5,0.847)};
    \addplot[color6,mark=pentagon*,line width=1pt] coordinates {(1,0.434) (2,0.436) (3,0.610) (4,0.245) (5,0.915)};
    \addplot[black,mark=o,line width=1pt] coordinates {(1,0.162) (2,0.198) (3,0.216) (4,0.137) (5,0.494)};
    \addplot[magenta!70!black,mark=star,line width=1pt] coordinates {(1,0.649) (2,0.692) (3,0.803) (4,0.801) (5,0.945)};
\end{axis}
}

\footnotesize{
\begin{axis}[
at={(0cm,-3.0cm)},
    ymajorgrids,
    xmajorgrids,
    grid style=dashed,
    width=0.29\textwidth, height=0.23\textwidth,
    legend columns=-1,
    xmin=0.8, xmax=5.2,
    ymin=0.2, ymax=1.2,
    xtick={1,2,3,4,5},
    xticklabels={\scriptsize{2},\scriptsize{3},\scriptsize{4},\scriptsize{5-6},\scriptsize{7+}},
    xlabel=\footnotesize{cpWER($\downarrow$)},
    xlabel style={yshift=0.1em,align=center},
    scaled ticks=false,
    every axis plot/.append style={mark options={fill opacity=0.5,draw opacity=1}},
    ]
    \addplot[color5,mark=*,line width=1pt] coordinates {(1,0.325) (2,0.371) (3,0.501) (4,0.423) (5,0.784)};
    \addplot[color3,mark=square*,line width=1pt] coordinates {(1,0.387) (2,0.382) (3,0.367) (4,0.444) (5,1.113)};
    \addplot[color2,mark=triangle*,line width=1pt] coordinates {(1,0.484) (2,0.317) (3,0.529) (4,0.323) (5,0.617)};
    \addplot[color4,mark=diamond*,line width=1pt] coordinates {(1,0.337) (2,0.452) (3,0.541) (4,0.557) (5,0.633)};
    \addplot[color6,mark=pentagon*,line width=1pt] coordinates {(1,0.523) (2,0.426) (3,0.567) (4,0.444) (5,0.631)};
    \addplot[black,mark=o,line width=1pt] coordinates {(1,0.283) (2,0.366) (3,0.449) (4,0.259) (5,0.599)};
    \addplot[magenta!70!black,mark=star,line width=1pt] coordinates {(1,0.705) (2,0.716) (3,0.801) (4,0.831) (5,0.909)};
\end{axis}
}
\footnotesize{
\begin{axis}[
at={(3.8cm,-3.0cm)},
    ymajorgrids,
    xmajorgrids,
    grid style=dashed,
    width=0.29\textwidth, height=0.23\textwidth,
    legend columns=-1,
    xmin=0.8, xmax=5.2,
    ymin=0.3, ymax=1.8,
    xtick={1,2,3,4,5},
    xticklabels={\scriptsize{2},\scriptsize{3},\scriptsize{4},\scriptsize{5-6},\scriptsize{7+}},
    xlabel=\footnotesize{tcpWER($\downarrow$)},
    xlabel style={yshift=0.1em,align=center},
    scaled ticks=false,
    every axis plot/.append style={mark options={fill opacity=0.5,draw opacity=1}},
    ]
    \addplot[color5,mark=*,line width=1pt] coordinates {(1,1.358) (2,1.276) (3,1.365) (4,1.358) (5,1.255)};
    \addplot[color3,mark=square*,line width=1pt] coordinates {(1,1.012) (2,1.078) (3,1.000) (4,1.513) (5,1.478)};
    \addplot[color2,mark=triangle*,line width=1pt] coordinates {(1,1.181) (2,1.178) (3,1.206) (4,1.702) (5,1.502)};
    \addplot[color4,mark=diamond*,line width=1pt] coordinates {(1,0.958) (2,0.913) (3,1.014) (4,1.110) (5,1.310)};
    \addplot[color6,mark=pentagon*,line width=1pt] coordinates {(1,0.773) (2,0.876) (3,0.933) (4,0.709) (5,0.972)};
    \addplot[black,mark=o,line width=1pt] coordinates {(1,0.352) (2,0.444) (3,0.617) (4,0.375) (5,0.712)};
    \addplot[magenta!70!black,mark=star,line width=1pt] coordinates {(1,1.025) (2,1.055) (3,1.145) (4,1.071) (5,1.093)};
\end{axis}
}
\end{tikzpicture}

\caption{Performance across speaker counts on \name. Omni LLM performance generally degrades with increasing speaker numbers.}
\label{fig:speaker}
\end{figure}
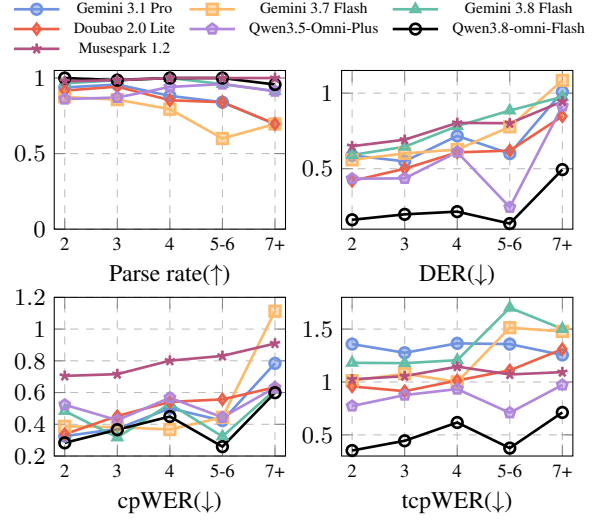

\subsection{Analysis of Speaker Number}

\definecolor{color4}{RGB}{235,096,070}
\definecolor{color5}{RGB}{111,146,225}
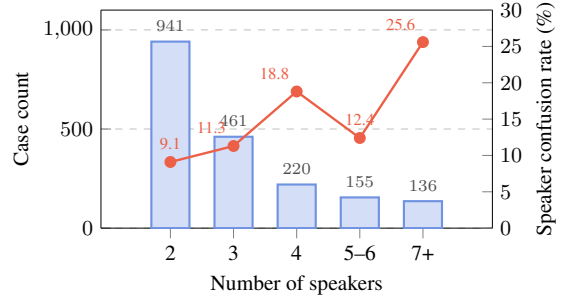
\begin{figure}[t]
\centering
\resizebox{0.46\textwidth}{!}{%
\begin{tikzpicture}
\footnotesize
\begin{axis}[
    at={(0cm,0cm)},
    width=0.46\textwidth, height=0.30\textwidth,
    ybar,
    bar width=16pt,
    axis y line*=left,
    ymajorgrids, grid style=dashed,
    xmin=0.5, xmax=5.5,
    ymin=0, ymax=1100,
    xtick={1,2,3,4,5},
    xtick pos=bottom,
    xticklabels={2,3,4,5--6,7+},
    xlabel=\footnotesize{Number of speakers},
    ylabel=\footnotesize{Case count},
    scaled ticks=false,
    enlarge x limits=0.12,
    nodes near coords,
    nodes near coords style={font=\scriptsize,color=black!70},
    every node near coord/.append style={yshift=1pt},
    ]
    \addplot[fill=color5!30,draw=color5,line width=0.9pt] coordinates {(1,941) (2,461) (3,220) (4,155) (5,136)};
\end{axis}
\begin{axis}[
    at={(0cm,0cm)},
    width=0.46\textwidth, height=0.30\textwidth,
    axis y line*=right,
    axis x line=none,
    xmin=0.5, xmax=5.5,
    ymin=0, ymax=30,
    ytick={0,5,10,15,20,25,30},
    ylabel=\footnotesize{Speaker confusion rate (\%)},
    scaled ticks=false,
    enlarge x limits=0.12,
    ]
    \addplot[color4,mark=*,line width=1pt,mark options={fill=color4}]
        coordinates {(1,9.1) (2,11.3) (3,18.8) (4,12.4) (5,25.6)};
    \node[font=\scriptsize,color=color4,anchor=south,yshift=2pt] at (axis cs:1,9.1) {9.1};
    \node[font=\scriptsize,color=color4,anchor=south east,yshift=2pt] at (axis cs:2,11.3) {11.3};
    \node[font=\scriptsize,color=color4,anchor=south east,yshift=2pt] at (axis cs:3,18.8) {18.8};
    \node[font=\scriptsize,color=color4,anchor=south,yshift=2pt] at (axis cs:4,12.4) {12.4};
    \node[font=\scriptsize,color=color4,anchor=south east,yshift=2pt] at (axis cs:5,25.6) {25.6};
\end{axis}
\end{tikzpicture}%
}
\caption{Analysis of speaker identification. Blue bars represents the quantity of cases per category, and the red line gives the speaker confusion rate.}
\label{fig:speaker-confusion}
\end{figure}

Disentangling multiple speakers remains a non-trivial challenge for omni-LLMs. In this section, we investigate the impact of speaker count on MsASR performance. While \name contains up to 20 speakers, we partition the speaker counts into five categories as shown on right side of Fig \ref{fig:category}, and exclude a single-speaker outlier to maintain a balanced sample distribution.

\definecolor{color2}{RGB}{098,190,166}
\definecolor{color3}{RGB}{253,186,107}
\definecolor{color4}{RGB}{235,096,070}
\definecolor{deltagreen}{RGB}{046,125,050}
\definecolor{deltared}{RGB}{198,040,040}

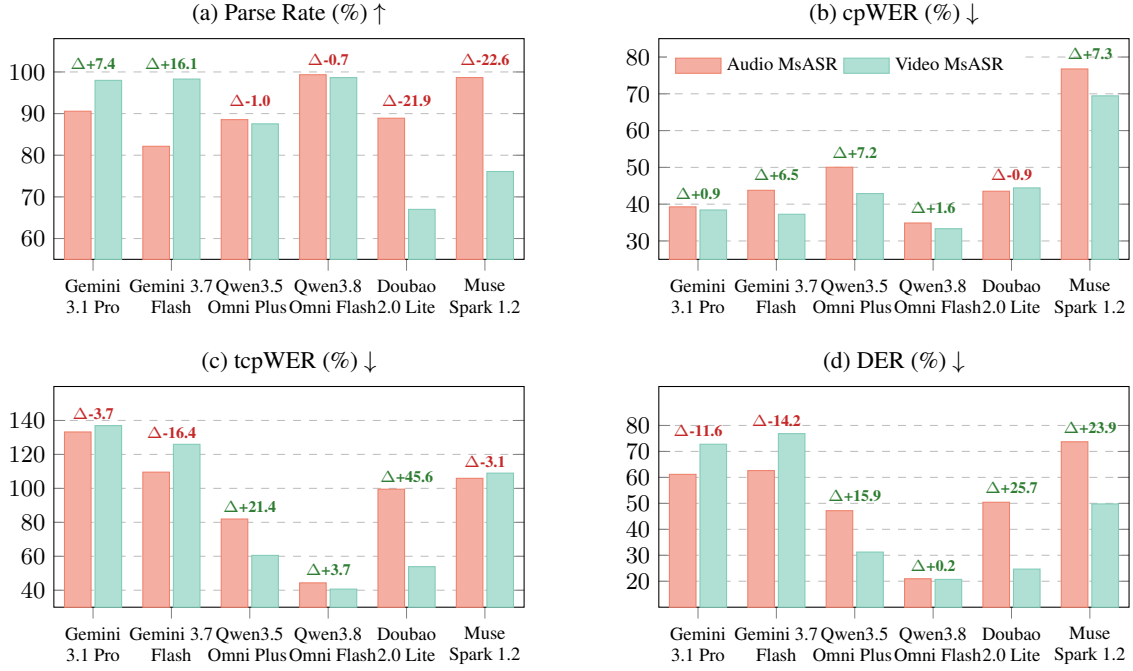
\begin{figure*}[t]
\centering
\begin{tikzpicture}
\footnotesize{
\begin{axis}[
    at={(0,0)},
    ymajorgrids,
    grid style=dashed,
    width=7.8cm,
    height=4.5cm,
    ybar=1.5pt,
    enlarge x limits=0.10,
    bar width=10pt,
    symbolic x coords={M1,M2,M3,M4,M5,M6},
    xtick=data,
    xtick pos=bottom,
    xticklabels={
        {Gemini\\3.1 Pro},
        {Gemini 3.7\\Flash},
        {Qwen3.5\\Omni Plus~~},
        {Qwen3.8\\~~Omni Flash},
        {Doubao\\2.0 Lite},
        {Muse\\Spark 1.2}
    },
    xticklabel style={align=center, font=\scriptsize},
    title={\footnotesize{(a) Parse Rate (\%) $\uparrow$}},
    title style={yshift=-4pt},
    ymin=55, ymax=108,
    ytick={60,70,80,90,100},
    scaled ticks=false,
]
    \addplot[fill=color4!50, draw=color4!80, area legend]
        coordinates {(M1,90.57)(M2,82.15)(M3,88.55)
                     (M4,99.33)(M5,88.89)(M6,98.65)};
    \addplot[fill=color2!50, draw=color2!80, area legend]
        coordinates {(M1,97.98)(M2,98.29)(M3,87.54)
                     (M4,98.65)(M5,67.00)(M6,76.09)};
    \node[font=\tiny\bfseries, text=deltagreen] at (axis cs:M1,102) {$\Delta$+7.4};
    \node[font=\tiny\bfseries, text=deltagreen] at (axis cs:M2,102) {$\Delta$+16.1};
    \node[font=\tiny\bfseries, text=deltared]   at (axis cs:M3,93)  {$\Delta${-}1.0};
    \node[font=\tiny\bfseries, text=deltared]   at (axis cs:M4,103) {$\Delta${-}0.7};
    \node[font=\tiny\bfseries, text=deltared]   at (axis cs:M5,93)  {$\Delta${-}21.9};
    \node[font=\tiny\bfseries, text=deltared]   at (axis cs:M6,103) {$\Delta${-}22.6};
\end{axis}
}
\footnotesize{
\begin{axis}[
    at={(8.0cm,0)},
    ymajorgrids,
    grid style=dashed,
    width=7.8cm,
    height=4.5cm,
    ybar=1.5pt,
    enlarge x limits=0.10,
    bar width=10pt,
    symbolic x coords={M1,M2,M3,M4,M5,M6},
    xtick=data,
    xtick pos=bottom,
    xticklabels={
        {Gemini\\3.1 Pro},
        {Gemini 3.7\\Flash},
        {Qwen3.5\\Omni Plus~~},
        {Qwen3.8\\~~Omni Flash},
        {Doubao\\2.0 Lite},
        {Muse\\Spark 1.2}
    },
    xticklabel style={align=center, font=\scriptsize},
    title={\footnotesize{(b) cpWER (\%) $\downarrow$}},
    title style={yshift=-4pt},
    ymin=25, ymax=85,
    ytick={30,40,50,60,70,80},
    scaled ticks=false,
    legend style={
        at={(0.39,0.79)},
        anchor=south,
        font=\scriptsize,
        draw=none,
        legend columns=-1,
    },
]
    \addplot[fill=color4!50, draw=color4!80, area legend]
        coordinates {(M1,39.27)(M2,43.77)(M3,50.04)
                     (M4,34.88)(M5,43.52)(M6,76.76)};
    \addplot[fill=color2!50, draw=color2!80, area legend]
        coordinates {(M1,38.42)(M2,37.26)(M3,42.88)
                     (M4,33.33)(M5,44.42)(M6,69.45)};
    \legend{Audio MsASR, Video MsASR}
    \node[font=\tiny\bfseries, text=deltagreen] at (axis cs:M1,43)  {$\Delta$+0.9};
    \node[font=\tiny\bfseries, text=deltagreen] at (axis cs:M2,48)  {$\Delta$+6.5};
    \node[font=\tiny\bfseries, text=deltagreen] at (axis cs:M3,54)  {$\Delta$+7.2};
    \node[font=\tiny\bfseries, text=deltagreen] at (axis cs:M4,39)  {$\Delta$+1.6};
    \node[font=\tiny\bfseries, text=deltared]   at (axis cs:M5,48)  {$\Delta${-}0.9};
    \node[font=\tiny\bfseries, text=deltagreen] at (axis cs:M6,81)  {$\Delta$+7.3};
\end{axis}
}
\footnotesize{
\begin{axis}[
    at={(0,-4.6cm)},
    ymajorgrids,
    grid style=dashed,
    width=7.8cm,
    height=4.5cm,
    ybar=1.5pt,
    enlarge x limits=0.10,
    bar width=10pt,
    symbolic x coords={M1,M2,M3,M4,M5,M6},
    xtick=data,
    xtick pos=bottom,
    xticklabels={
        {Gemini\\3.1 Pro},
        {Gemini 3.7\\Flash},
        {Qwen3.5\\Omni Plus~~},
        {Qwen3.8\\~~Omni Flash},
        {Doubao\\2.0 Lite},
        {Muse\\Spark 1.2}
    },
    xticklabel style={align=center, font=\scriptsize},
    title={\footnotesize{(c) tcpWER (\%) $\downarrow$}},
    title style={yshift=-4pt},
    ymin=30, ymax=160,
    ytick={40,60,80,100,120,140},
    scaled ticks=false,
]
    \addplot[fill=color4!50, draw=color4!80, area legend]
        coordinates {(M1,133.22)(M2,109.56)(M3,81.94)
                     (M4,44.34)(M5,99.51)(M6,105.92)};
    \addplot[fill=color2!50, draw=color2!80, area legend]
        coordinates {(M1,136.96)(M2,125.95)(M3,60.53)
                     (M4,40.66)(M5,53.89)(M6,108.98)};
    \node[font=\tiny\bfseries, text=deltared]   at (axis cs:M1,144) {$\Delta${-}3.7};
    \node[font=\tiny\bfseries, text=deltared]   at (axis cs:M2,133) {$\Delta${-}16.4};
    \node[font=\tiny\bfseries, text=deltagreen] at (axis cs:M3,89)  {$\Delta$+21.4};
    \node[font=\tiny\bfseries, text=deltagreen] at (axis cs:M4,52)  {$\Delta$+3.7};
    \node[font=\tiny\bfseries, text=deltagreen] at (axis cs:M5,107) {$\Delta$+45.6};
    \node[font=\tiny\bfseries, text=deltared]   at (axis cs:M6,116) {$\Delta${-}3.1};
\end{axis}
}
\footnotesize{
\begin{axis}[
    at={(8.0cm,-4.6cm)},
    ymajorgrids,
    grid style=dashed,
    width=7.8cm,
    height=4.5cm,
    ybar=1.5pt,
    enlarge x limits=0.10,
    bar width=10pt,
    symbolic x coords={M1,M2,M3,M4,M5,M6},
    xtick=data,
    xtick pos=bottom,
    xticklabels={
        {Gemini\\3.1 Pro},
        {Gemini 3.7\\Flash},
        {Qwen3.5\\Omni Plus~~},
        {Qwen3.8\\~~Omni Flash},
        {Doubao\\2.0 Lite},
        {Muse\\Spark 1.2}
    },
    xticklabel style={align=center, font=\scriptsize},
    title={\footnotesize{(d) DER (\%) $\downarrow$}},
    title style={yshift=-4pt},
    ymin=10, ymax=95,
    ytick={20,30,40,50,60,70,80},
    scaled ticks=false,
]
    \addplot[fill=color4!50, draw=color4!80, area legend]
        coordinates {(M1,61.17)(M2,62.62)(M3,47.17)
                     (M4,20.96)(M5,50.42)(M6,73.72)};
    \addplot[fill=color2!50, draw=color2!80, area legend]
        coordinates {(M1,72.77)(M2,76.85)(M3,31.24)
                     (M4,20.72)(M5,24.68)(M6,49.78)};
    \node[font=\tiny\bfseries, text=deltared]   at (axis cs:M1,78)  {$\Delta${-}11.6};
    \node[font=\tiny\bfseries, text=deltared]   at (axis cs:M2,82)  {$\Delta${-}14.2};
    \node[font=\tiny\bfseries, text=deltagreen] at (axis cs:M3,53)  {$\Delta$+15.9};
    \node[font=\tiny\bfseries, text=deltagreen] at (axis cs:M4,26)  {$\Delta$+0.2};
    \node[font=\tiny\bfseries, text=deltagreen] at (axis cs:M5,56)  {$\Delta$+25.7};
    \node[font=\tiny\bfseries, text=deltagreen] at (axis cs:M6,79)  {$\Delta$+23.9};
\end{axis}
}
\end{tikzpicture}
\caption{Comparison between audio and video MsASR, where $\Delta$ represents the advantage of visual modality.}
\label{fig:video}
\end{figure*}

As illustrated in Fig. \ref{fig:speaker}, the parse rates of omni LLMs remain largely stable, suggesting that their instruction-following capability does not degrade with acoustic complexity. 
However, metrics factoring in speaker identification (DER, cpWER, and tcpWER) deteriorate substantially as the number of speakers grows. Notably, the cpWER of \texttt{Gemini 3.7 Flash} surges from $0.387$ with two speakers to $1.113$ in the most crowded setting. 

Fig. \ref{fig:speaker-confusion} further reports the speaker confusion rate: 
\begin{equation}
    \text{SCR}=\frac{T_{\text{conf}}}{T_{\text{total}}}
\end{equation}
Experimental results reveal a pronounced upward trajectory despite a slight fluctuation at five speakers. These findings highlight that \textit{robust speaker identification in multi-speaker scenarios} remains an open challenge for omni-modal LLMs.

\subsection{Multi-speaker Video ASR}

Prior benchmarks predominantly overlook the visual modality's potential to facilitate speaker identification and transcription. Human perception naturally integrates visual and acoustic cues: listeners direct their gaze toward the sound source to disambiguate speakers under acoustic ambiguity. 
Motivated by this, \name establish a parallel evaluation comparing audio-only and audio-visual (video) MsASR. We investigate whether current omni-modal LLMs have passed the inflection point where visual context effectively enhances MsASR, or whether redundant visual signals conversely introduce confounding noise.

As illustrated in Fig. \ref{fig:video}, \textit{the visual modality does not consistently benefit MsASR}. 
In terms of parsing rate, neither modality exhibits a uniform advantage: the Qwen family shows near-identical adherence across settings, the Gemini family yields higher parsing rates with video inputs, whereas other models perform better with audio. 
This discrepancy reflects a latent modality gap and highlights divergent cross-modal alignment strategies across models.
Notably, \texttt{Qwen3.8-omni-flash} demonstrates high cross-modal consistency across both settings. 
Intriguingly, while \texttt{Doubao-seed-2.0-lite} achieves comparable cpWER between modalities, its video MsASR substantially outperforms the audio counterpart in tcpWER and DER, indicating that visual cues significantly aid Doubao's temporal boundary and timestamp prediction. Conversely, \texttt{Gemini-3.7} exhibits an inverted trend: video context improves cpWER but degrades tcpWER and DER, suggesting that visual signals hinder Gemini's temporal grounding capabilities.
Further DER analysis shows that \textit{video inputs aid speech activity detection}, reducing the speaker miss rate by 0.15, \textit{while failing to improve speaker discrimination}, resulting in a 0.014 increase in speaker confusion.

Experimental findings of this paper highlight \textit{fine-grained temporal localization}, \textit{speaker disambiguation}, and \textit{cross-modal synergy} as key challenges for future omni-modal foundation models.

\section{Conclusions}

We present \name, a multimodal, multilingual, and long-form benchmark for multi-speaker ASR of omni LLMs. It comprises 297 instances with parallel audio and video inputs across six languages. Further, we conduct extensive experiments on frontier omni LLMs, revealing that temporal grounding and speaker disambiguation remain key bottlenecks. We hope \name will facilitate progress in temporal grounding, speaker disambiguation, and audio-visual integration for omni-modal LLMs.

\section*{Limitations}
The pseudo-labels of \name are generated by \texttt{MOSS Transcribe Diarize} and \texttt{Gemini 3.1 Pro}, and we notice that timestamps are corrected during expert verification only when obvious errors are found. This may bias the evaluation in favor of \texttt{MOSS Transcribe Diarize}, particularly on time-sensitive metrics. Additionally, inference with commercial APIs is stochastic, so the exact reproducibility of the \name benchmark results cannot be guaranteed.

\section*{Ethical Considerations}

\paragraph{Data source and release.} All videos are publicly available on YouTube, and \textsc{Omnilingual-MsASR} is released for non-commercial research only, in accordance with the platform's terms of service. Speaker labels are anonymous indices local to each video and are not linked to real-world identities. We will remove any instance upon request from the content owner or a depicted individual.

\paragraph{Privacy.} Facial cues are used only to link speakers within the same video during annotation, and face crops are not part of the released benchmark. A dedicated privacy audit identified three French instances containing personally identifiable information, namely real social media handles and email addresses, and these instances were removed.

\paragraph{Annotators.} The verification was carried out by professional linguistic experts who were informed of the research purpose and fairly compensated through the company. No personal information about annotators was collected.

\paragraph{Potential misuse.} Speaker-attributed transcription of long-form video could be misused for surveillance or for profiling individuals. \textsc{Omnilingual-MsASR} is designed solely to evaluate model capabilities, and we discourage its use for identifying or tracking specific people.




\bibliography{custom}

\appendix

\definecolor{PromptBoxTitleColor}{RGB}{76,76,76}
\definecolor{PromptBoxColor}{RGB}{247,247,255}
\definecolor{block-gray}{gray}{0.85}

\newtcolorbox{mybackground}{colback=block-gray!50,grow to right by=0mm,grow to left by=0mm,boxrule=0pt,boxsep=0pt,breakable}

\section{Specifics Document about Human Annotation Guidelines}
\label{append:guideline}

In this section, we introduce the annotation guidelines of linguistic expert annotation and review. The detailed instructions are shown below:


\newtcolorbox{quotebar}{
  empty,                     
  sharp corners,             
  boxrule=0pt,               
  leftrule=2.5pt,            
  colframe=gray!40,          
  left=8pt, right=0pt, top=2pt, bottom=2pt, 
  before skip=4pt,
  after skip=6pt
}

\newtcbox{\tagbox}[1][]{
  on line,
  arc=2pt,
  colback=gray!12,
  colframe=gray!30,
  boxrule=0.5pt,
  boxsep=1pt,
  left=3.5pt, right=3.5pt, top=1.5pt, bottom=1.5pt,
  fontupper=\small\ttfamily,
  #1
}

\newcommand{\sectiontitle}[1]{%
  \vspace{0.8em}\noindent{\Large\textbf{#1}}\par\vspace{0.4em}%
}

\setlength{\parindent}{0pt}
\setlength{\parskip}{4pt}

\paragraph{Languages:}~\\
Spanish, French, Japanese, Korean, Thai, Russian 

Each expert is responsible only for their assigned language.

\paragraph{Data Format:}~\\
\tagbox{\textcolor{red!80!black}{60}} data entries per language; each entry is on a separate sheet, as shown below.

\includegraphics[width=\linewidth]{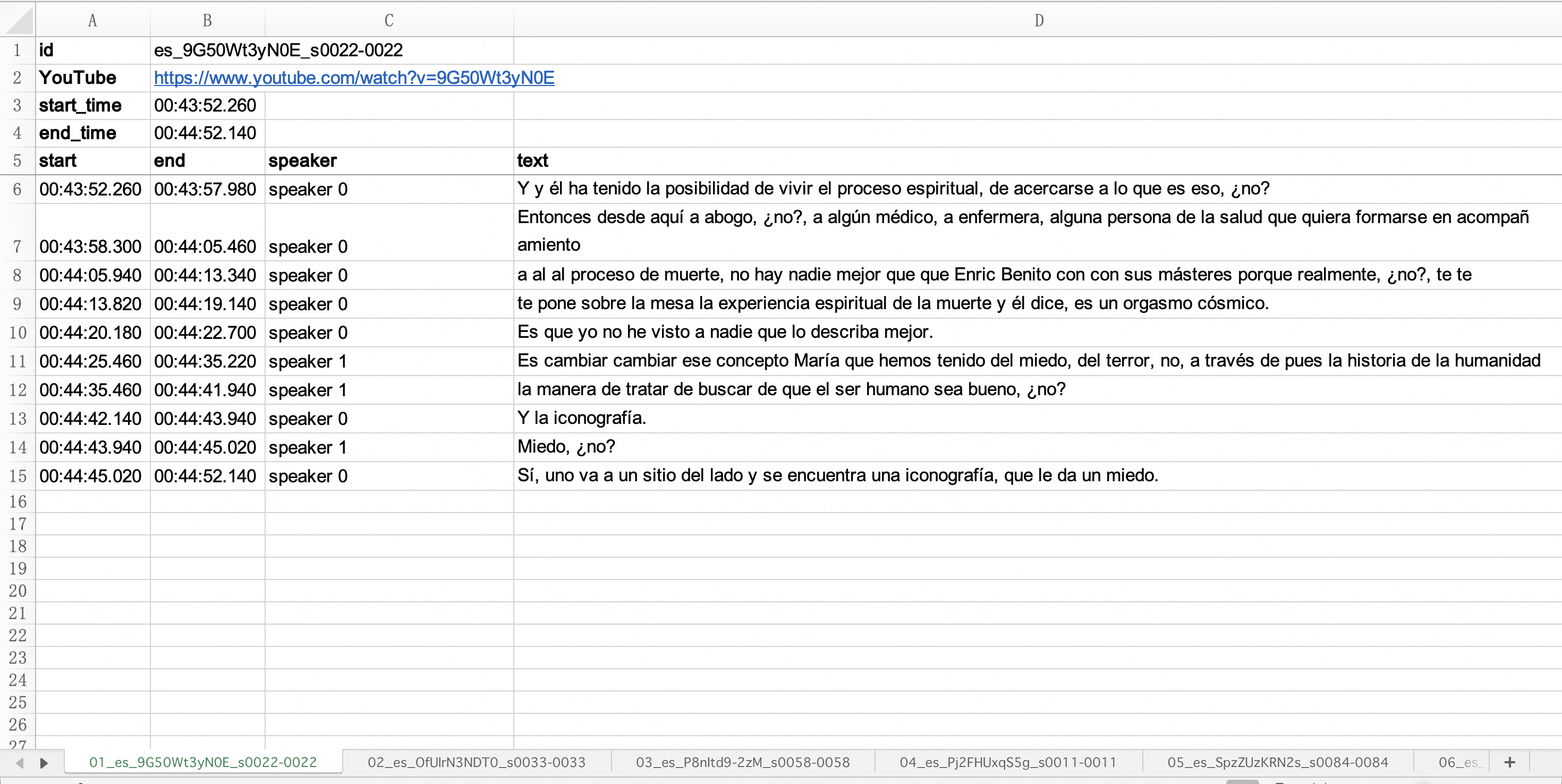}

\paragraph{Field Descriptions:}
\begin{itemize}[leftmargin=1.5em, itemsep=3pt, topsep=2pt]
  \item id: Task ID
  \item YouTube: Video URL
  \item start\_time / end\_time: Start and end timestamps of the \underline{video clip} corresponding to \tagbox{this entry}
  \item Starting from Row 6, \tagbox{each row} is a \tagbox{\textcolor{red!80!black}{speaker diarization record}}, including:
    \begin{itemize}[leftmargin=1.5em, itemsep=2pt, topsep=1pt, label=$\circ$]
      \item start: Speaker speech\textcolor{red!80!black}{start timestamp} 
      \item end: Speaker speech \textcolor{red!80!black}{end timestamp} 
      \item speaker: Speaker ID, 0-indexed
      \item text: \textcolor{red!80!black}{Transcribed content} of the speaker
      \item \textcolor{orange!90!black}{\faExclamationTriangle} Each record must contain only one speaker
    \end{itemize}
\end{itemize}

\paragraph{Task Description:}
\begin{enumerate}[leftmargin=1.5em, itemsep=5pt, topsep=2pt]
  \item Open the YouTube video link and skip forward to \tagbox{start\_time}.
  \item Verify whether the \textcolor{red!80!black}{speech activity} between \tagbox{start\_time} and \tagbox{end\_time} matches all \tagbox{speaker diarization records} (this requires verifying the correctness of \textcolor{red!80!black}{speaker identity, timestamps, and speech content}).
  \item Corrections (Add / Delete / Modify):
    \begin{enumerate}[leftmargin=1.5em, itemsep=3pt, topsep=2pt, label=\alph*.]
      \item If the timestamps, speaker, or speech content are incorrect, modify them directly in place.
      \item If a single \tagbox{\textcolor{red!80!black}{speaker diarization record}} actually contains multiple speakers, it must be split into separate records. e.g., If the segment in Row 10 actually has 3 speakers (speaker 0, speaker 1, speaker 2) speaking sequentially, insert two rows below Row 10. These three rows should separately record each speaker's start and end times (\texttt{start}, \texttt{end}), label the speaker with the corresponding ID, and transcribe the actual spoken text.
      \item If an annotator identifies an unindexed speaker, \textcolor{red!80!black}{simply add a new speaker ID}. e.g., If there are actually three distinct speakers in the clip above, use \texttt{speaker 2} where necessary.
      \item \tagbox{Timestamps} must be annotated to \textcolor{red!80!black}{one decimal place}.
    \end{enumerate}
\end{enumerate}

\paragraph{Notes:}
\begin{itemize}[leftmargin=1.2em, itemsep=2pt, topsep=1pt]
  \item If the video begins with \textbf{group singing/chorus}, treat the ensemble as a \tagbox{\textcolor{red!80!black}{single speaker}}.
  \item If one or two individuals speak first followed by group singing (where the transitions are clearly audible), each speaker and the subsequent singing must be \textbf{segmented and labeled separately}.
\end{itemize}

\section{Inference Prompt for MsASR}
\label{appendix:prompt}
The prompt configuration of \name utilized for omni-LLMs inference is detailed below. 

\begin{mybackground}
\# Role \\
You are a professional audio transcription expert specializing in Automatic Speech Recognition (ASR) with Speaker Attribution (multi-speaker labels and timestamps).\\ 

\#  Task \\
Listen to the given audio, then generate ASR text with speaker labels and timestamps. Use the acoustic cues (voice timbre, pitch, speaking rate and style) only to tell the speakers apart and keep their IDs consistent; the transcribed text itself must come from what is spoken.\\

\#  Output Format Rules \\
Please strictly follow the XML-like tag format below. Do not include any additional explanations, Markdown code block markers (e.g., \textasciigrave\textasciigrave\textasciigrave xml), or idle chatter—output only the final result string.\\

1. **Overall structure**: \\
    \quad - Start with \textasciigrave<soc>\textasciigrave. \\
    \quad - End with \textasciigrave<eoc>\textasciigrave.\\

2. **Segment structure**: \\
Each utterance segment must be concatenated in exactly the following order: \\
\textasciigrave<sos>\textasciigrave + \textasciigrave<start time>\textasciigrave + \textasciigrave ASR text content\textasciigrave + \textasciigrave<end time>\textasciigrave + \textasciigrave<speaker ID>\textasciigrave + \textasciigrave<eos>\textasciigrave\\

Where: \\
- \textasciigrave<sos>\textasciigrave: segment start marker \\
- \textasciigrave<start time>\textasciigrave: floating-point format, **must keep exactly two decimal places** (e.g., 0.00, 12.50). \\
- \textasciigrave ASR text content\textasciigrave: the recognized text. \\
- \textasciigrave<end time>\textasciigrave: floating-point format, **must keep exactly two decimal places**. \\
- \textasciigrave<speaker ID>\textasciigrave: formatted as \textasciigrave<speakerN>\textasciigrave, where N is an integer. \\
- \textasciigrave<eos>\textasciigrave: segment end marker\\

3. **Speaker ID assignment logic**: \\
- IDs start counting from 0. \\
- Assign IDs sequentially according to the order of each speaker's **first appearance** in the audio. \\
- In later segments, the same speaker must reuse the already assigned ID. \\
- When a new speaker appears, assign the next available integer ID.\\

\#  Few-Shot Example \\
The input audio contains two speakers: \\
- Speaker A speaks ``Mm.'' from 0.00s to 0.67s \\
- Speaker B speaks ``Probably more towards three than two.'' from 1.68s to 2.81s \\

The correct output is: \\
<soc><sos><0.00>Mm.<0.67><speaker0><eos> <sos><1.68>Probably more towards three than two.<2.81><speaker1><eos><eoc>\\

Now please process the input audio according to the requirements above \\
Transcribe the dialogue with speaker identification and timestamps. \\
Output format: \\
<soc><sos><start\_time>text<end\_time> <speakerX><eos>...<eoc>.

\end{mybackground}

For video MsASR, we utilize the same prompt, except that \textit{"audio"} is replaced with \textit{"video"}. Note that the inference settings of \texttt{MOSS-Transcribe-Diarize-0.9B} are consistent with the official implementation, as specialized models do not require task-specific prompts. Similarly, the output of MOSS is parsed following the default format, in contrast to Sec. \ref{tab:format}.

\section{Data Statement and License}
\paragraph{Data Usage Statement.} For third-party original audio/video, this dataset provides only index references and text annotation results. 
It does not include, copy or redistribute any such original audio/video content. It is for academic research only, not for commercial use, and the copyright of such original audio/video content belongs to its respective creators and rights holders. When accessing and using the referenced material, users must comply with the terms of service of the original content platforms.

\paragraph{License.} The above indexes and text annotations are licensed under CC BY-NC 4.0 (Attribution–NonCommercial 4.0 International). License text: \url{https://creativecommons.org/licenses/by-nc/4.0/legalcode.en}. This license applies only to the indexes and annotations provided by this repository and does not cover any rights in the original audio/video content.

\paragraph{Offensive content.} During the benchmark build process, we discovered that some of the data contained profanity and other offensive content. We removed the data containing offensive content.

\end{document}